\documentclass{article}

\usepackage[preprint]{nips_2018}

\usepackage[utf8]{inputenc} 
\usepackage[T1]{fontenc}    
\usepackage{hyperref}       
\usepackage{url}            
\usepackage{booktabs}       
\usepackage{amsfonts}       
\usepackage{nicefrac}       
\usepackage{microtype}      
\usepackage{graphicx}
\usepackage{algorithm}
\usepackage{algorithmic}
\usepackage{makecell}
\usepackage{mathtools}

\usepackage[backend=biber]{biblatex}
\title{AugLabel: Exploiting Word Representations to Augment Labels for Face Attribute Classification}

\author{%
  Binod Bhattarai\thanks{Equal Contribution} \\
  Imperial College London\\
  \texttt{b.bhattarai@imperial.ac.uk} \\
  \And
   Rumeysa Bodur \thanks{Equal Contribution} \\
  Imperial College London \\
  \texttt{r.bodur18@imperial.ac.uk} \\
  \AND
  Tae-Kyun Kim \\
  Imperial College London \\
  \texttt{tk.kim@imperial.ac.uk}
}

\begin{document}

\def\eg{\emph{e.g}\bmvaOneDot}
\def\Eg{\emph{E.g}\bmvaOneDot}
\def\etal{\emph{et al}\bmvaOneDot}


\maketitle

\def\etal{et al\onedot}
\def\etc{etc\onedot}
\def\ie{i.e\onedot}
\def\eg{e.g\onedot}
\def\cf{cf\onedot}
\def\vs{vs\onedot}
\def\grad{\nabla}
\def\b{\textbf{b}}
\def\v{\textbf{v}}
\def\a{\boldsymbol{\alpha}}
\def\sign{\textrm{sign}}
\def\pd{\partial}
\def\T{\mathcal{T}}
\def\R{\mathbb{R}}
\def\Reg{\mathcal{R}}
\def\X{\mathcal{X}}
\def\I{\mathcal{I}}
\def\F{\mathcal{F}}
\def\Obj{\mathcal{O}}
\def\V{\mathcal{V}}
\def\W{\textbf{W}}
\def\w{\textbf{w}}
\def\z{\textbf{z}}
\def\h{\textbf{h}}
\def\c{\textbf{c}}
\def\1{\textbf{1}}
\def\x{\textbf{x}}
\def\c{\textbf{c}}
\def\s{\textbf{s}}
\def\d{\boldsymbol{\delta}}
\def\y{\textbf{y}}
\def\l{\textbf{l}}
\def\ock{$1$-call@$K$}
\def\oc{$1$-call@}
\def\TODO{\textcolor{red}{TODO} }

\begin{abstract}
Augmenting data in image space (eg.~flipping, cropping etc) and activation space (eg. dropout) are being widely used to regularise deep neural networks and have been successfully applied on several computer vision tasks. Unlike previous works, which are mostly focused on doing augmentation in the aforementioned domains, we propose to do augmentation in label space. 
In this paper, we present a novel method to generate fixed dimensional labels with continuous values for images by exploiting the \emph{word2vec} representations of the existing categorical labels. We then append these representations with existing categorical labels and train the model. 
We validated our idea on two challenging face attribute classification data sets viz. CelebA and LFWA. Our extensive experiments show that the augmented labels improve the performance of the competitive deep learning baseline and reduce the need of annotated real data up to 50\%, while attaining a performance similar to the state-of-the-art methods. 
\end{abstract}
\section{Introduction}
\label{intro}
Face attribute recognition~\cite{bhattarai2016deep,kumar2008facetracer, kalayeh2017improving, hand2017attributes} is a popular and an important 
research topic in the field of face image analysis. Face attributes include higher order features of face, such as \emph{smiling, sad,  happy, male, wearing glass}, which are intuitive and easily understandable. Such features have been successfully used in other face analysis tasks such as 
face verification and recognition~\cite{kumar2009attribute} and face retrieval~\cite{kumar2011describable}.
In addition to this, the recognition of face attributes, such as \emph{smiling, sad, happy etc.}, has been used in business intelligence and human computer interaction.

With the advent of deep learning algorithms~\cite{krizhevsky2012imagenet}, the performance of face attribute recognition models has improved by a large margin~\cite{liu2015deep}.
Despite this improvement, deep learning frameworks are still prone to overfitting. To overcome such issues, several existing 
works such as data augmentation~\cite{krizhevsky2012imagenet} and dropout~\cite{srivastava2014dropout}~ are being successfully used.
These works mostly focus on input space and the intermediate steps of the framework. However, attention to regularise in output space is only recently given. 
DisturbLabel~\cite{xie2016disturblabel} proposes to randomly flip the label to augment in output space. Similarly, ~\cite{pereyra2017regularizing} proposes to
regularise the output space by penalising the low entropy output 
distributions. However, these papers rely only on existing categorical 
labels. In this paper, we propose a complementary method to existing 
methods for regularising in output space by augmenting categorical labels with continuous labels. To this end, we propose to utilise the semantic representations of categorical
labels to generate continuous labels for the training images. Semantic representations of categorical labels are computed from the \emph{word2vec}~\cite{mikolov2013distributed, bengio2003neural} representations, which are continuous vector representations of words derived from a large corpus of text. These representations 
carry both semantic and syntactic representations of words and are so mathematical that they can do representations like
\emph{France + Paris - England $\approx$  London}. We leverage this strength of word2vec representations to generate new labels for face attribute
classification. Fig.~\ref{fig:proposed_method} shows the pipeline of the proposed method. To the best of our knowledge, this is the first work to utilise such representations to generate labels, augment them in label space and regularise the network. Moreover, this method is generic in nature and can be applied in addition to existing techniques in input 
space, intermediate layers and also in the output layers. Our contributions can be summarised in the following points:
\begin{itemize}
    \item A novel method is proposed to generate continuous valued labels from existing categorical labels leveraging \emph{word2vec}, semantic representations of the categorical labels.
    \item  Existing categorical labels are augmented with such representations to regularise the deep network.
    \item The proposed methods are generic and complementary to existing regularising techniques. 
    \item Evaluations are conducted on two competitive face attribute classification problems. Extensive experiments demonstrate that the existing method improves the performance over the existing regularising techniques and attains a performance similar to that of the state-of-the-art methods.
\end{itemize}

We organise the rest of the paper as follows. In Sec.~\ref{related_works}, we further discuss the related works. Then in Sec.~\ref{proposed_method} and Sec.~\ref{experiments},
we discuss our proposed method and the experiments we performed to evaluate our method in detail, respectively. Finally, we conclude our paper in Sec.~\ref{conclusions}.

\section{Related Works}
\label{related_works}
\noindent \textbf{Word2Vec.}
Word2vec~\cite{mikolov2013distributed,bengio2003neural} representations have been widely used in computer vision tasks, 
such as Zero Shot Learning~\cite{akata2013label, akata2015evaluation}, 
Visual Question Answering~\cite{malinowski2017ask}, 
conditional adversarial training~\cite{reed2016generative}, to mention but a few. However, we propose to use word embedding of categorical labels as an auxiliary task to regularise the main network in order to minimise the categorical loss.\\
\noindent \textbf{Multitask learning.} This learning framework is successfully applied on different computer vision problems~\cite{bhattarai2016cp, zhang2014facial,chen2017multi,xiong2015conditional}.
Our approach can be seen as one of the multi-task learning setups~\cite{caruana1997multitask}, where the parameters of both the main task (categorical loss) and the auxiliary task (continuous loss) are learned simultaneously. The major difference between our method and existing arts is that we do not need explicit annotations for the auxiliary task as we derive them from the categorical annotations using word2vec. \\
\noindent \textbf{Regularisation of Deep Networks.}
As we mentioned in the beginning, although the size of the annotated data to train deep networks is growing, deep learning setups still suffer from overfitting. This is more prevalent when access of annotated data is 
limited. To tackle this issue, data augmentation methods~\cite{krizhevsky2012imagenet} propose to add invariance 
in input space by generating synthetic examples by applying several 
geometric transformations such as translation, rotation, flipping and random cropping. GAN synthetic data~\cite{gecer2018eccv,baek2018augmented} is also being used to improve the generalisation
capability of the networks. Similarly, Dropout~\cite{srivastava2014dropout} proposes to add invariance on activation layers by randomly dropping out neurons. 
Recently, attention given to do augmentation in label space to regularise deep networks has also been increasing. 
DisturbLabel~\cite{xie2016disturblabel} proposes to do label augmentation 
by randomly corrupting some parts of the labels.~\cite{pereyra2017regularizing} proposes to penalise low entropy predictions. The limitations of these methods are that they solely rely on the given categorical labels. We propose to generate additional labels from existing categorical labels. Thus, our method can be seamlessly applied in addition to these existing techniques.\\ 
\noindent \textbf{Face Attributes Classification.}
Being one of the earliest works in face attribute classification, Face 
Tracer~\cite{kumar2008facetracer} computes Histogram of Gradients~\cite{dalal2005histograms} and color histograms to represent the faces and train SVM to make predictions. Similarly, ~\cite{liu2015deep} trains two CNNs end-to-end to localise the face and predict the attributes. MOON~\cite{rudd2016moon} proposes a mixed objective function to address the discrepancy of the distribution of attributes on source and target. MCNN~\cite{hand2017attributes}
proposes sharing parameters between related attributes in a multitask learning setup. Walk-and-Learn~\cite{wang2016walk} uses weather and location context obtained from egocentric video datasets to facilitate attribute prediction. AFFACT~\cite{gunther2017affact} proposes to do data augmentation in an alignment-free fashion using an ensemble of three ResNets for attribute classification. Kalayeh et al. ~\cite{kalayeh2017improving} utilises semantic segmentation to guide the learning process in order to improve facial attribute prediction. 
Sun et al.~\cite{sun2018bmvc} proposed hierarchical deep learning networks for learning parameters to make general to specific attribute predictions.
\section{Proposed method}
\label{proposed_method}
In this Section we discuss the proposed method in more detail. As
stated in the introduction, we are interested in generating continuous label representations from existing categorical labels. We augment existing labels with them and train the model with an extra level of generalisation added to the network. 
Categorical labels indicate the presence or absence of different 
attributes in a given face image. 

\begin{figure}
    \centering
    \includegraphics[width=1.0\textwidth, trim={0cm 5.1cm 2.5cm 2.7cm}, clip]{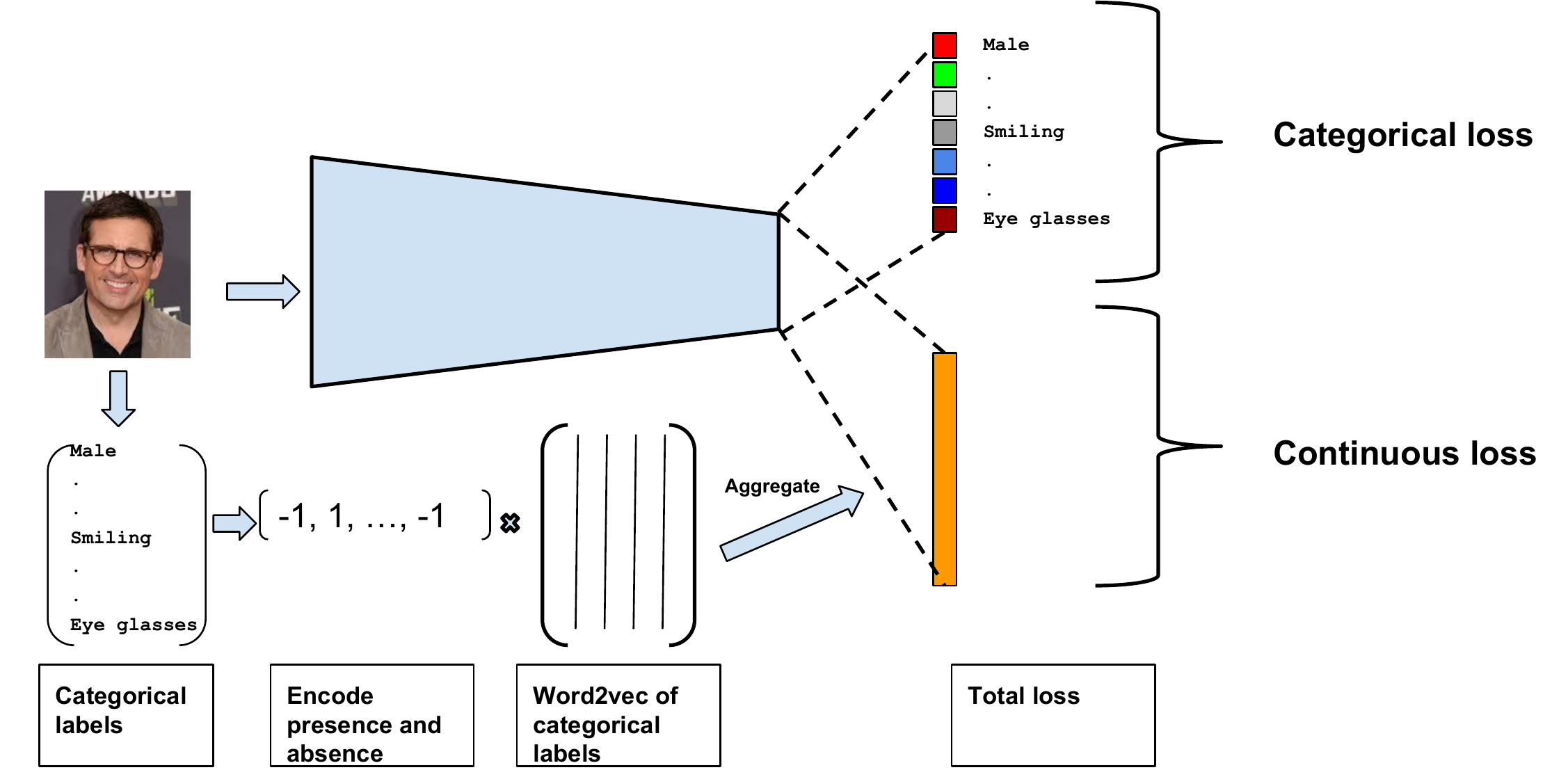}
    \caption{The schematic diagram of the proposed method pipeline. Textual descriptions of face attributes along with their presence/absence is utilised to generate 
    a compact continuous valued label. This new representations of labels are augmented along with existing categorical labels to train the CNN model.}
    \label{fig:proposed_method}
\end{figure}

We have a scenario where an input image $\x~\in \mathcal{R}^{(w \times h \times 3)}$  has its ground truth $\y \in \mathcal{R}^{m}$, where $m > 1$, which corresponds to 40 in our case. Out of 40 classes, one or more are 1 and the rest is 0, which indicates that more than one category can be present in an input image. Since we are evaluating our hypothesis on face attribute classification, the categories include different facial attributes such as \emph{gender, young, smiling, wearing glasses, wearing lipsticks, pointed noise} 
etc. At test time, for an input RGB image \x, we need to predict the presence or absence of multiple categories simultaneously. This kind of problems are well formulated as multi-label classification problems. To learn the parameters for making such predictions, a standard objective to minimise is the multiple binary cross-entropy loss.
Eqn.~\ref{eqn:binary_cross_entropy} formulates the standard (binary) cross entropy loss of the $j^{th}$ category. As stated before, $\y$ represents the ground truth vector and $\y^{p}$ is the predicted attribute vector. The limitations of this approach is that it considers every category equally different and independent from each other. However, in terms of 
face attributes, some of the attributes, such as \emph{young} and \emph{old} or \emph{female} and \emph{moustache} never co-occur. Similarly, some of the attributes are more closer than others. For example, \emph{male} and \emph{female} (gender) and \emph{smiling} and \emph{sad} (expressions) are closer than \emph{smiling} and \emph{wearing glasses} are. A recent attempt on sharing the parameters of related attributes~\cite{hand2017attributes} makes the model more robust than the standard baseline (not sharing parameters between the attributes). However, the main challenge with such an approach is the need of an expert to group the attributes. 
It also depends on the number of groups, and fine tuning is difficult and time consuming. Moreover, such hard assignment of an attribute to a group causes information loss, and assigning an attribute to a wrong group hurts the performance. 
To overcome such challenges and incorporate the relationship between the attributes, we propose to augment the existing labels with continuous valued labels and minimise both the categorical loss and continuous loss jointly. Please note, to generate such continuous labels, our method does not rely on any sophisticated external source of information. 
It only depends on the existing categorical labels and the textual semantic representations of the attributes.  We will describe the method to generate such representations shortly. 
Eqn.~\ref{eqn:obj_func} formulates the proposed objective, where the first part and the second part denotes the standard multiple binary cross-entropy loss (categorical loss) and augmented continuous loss respectively. In Eqn.~ $\alpha$ is the hyper-parameter that corresponds to the weights given to the two different losses. This value is determined by cross-validations.

\begin{equation}
    \mathcal{CE(\theta)} = -\y_{j}\log\y_{j}^{p} - (1- \y_{j})\log(1 - \y^{p}_{j}) 
\label{eqn:binary_cross_entropy}
\end{equation}

\noindent \textbf{Categorical Loss.} The first part of Eqn.~\ref{eqn:obj_func} represents the categorical loss, which is a standard multi-label classification formulation. 
In the Eqn. $N$ denotes the number of training examples, $\x$ is the training image with its ground truth label $\y$, and $\y^{p}$ denotes the predicted label. 
Unlike multi-class classification problems, labels are mutually exclusive and probabilities over all the labels sum to 1. We apply sigmoid on each output neuron, thus, it models the output of the network as as independent Bernoulli distribution per category. $j$ indicates the index of each category,
i'e attributes of the faces.\\

\noindent \textbf{Continuous Loss.} We propose to generate a compact and fixed dimension of $d$ continuous value filled labels from existing categorical labels by exploiting their 
word representations, which are commonly known as \emph{word2vec}~\cite{mikolov2013distributed}. These representations of words are learned from a large text corpus. As we mentioned in the introduction, these representations carry both syntactic and semantic meaning of the words. As a result, the relative relatedness between the words can be derived by computing Euclidean distance between their representations. For example, if we compute the Euclidean distance between \emph{black} and \emph{blonde} and \emph{black} and \emph{nose}, the distance between \emph{black} and \emph{blonde} will be smaller than the once between \emph{black} and \emph{nose}. As stated before, these representations support meaningful mathematical operations  such as addition and subtraction.
We leverage the power of these representations to synthesise new representations of annotated labels, and augment existing categorical labels to train the model. 

To generate continuous representations from categorical representations and their word representations, we first map all the attributes to their 
corresponding word representations using the pre-trained \emph{word2vec} models. For our experiments we use Glove~\cite{pennington2014glove}, publicly available representations of words~\footnote{https://github.com/stanfordnlp/GloVe}.
However, word representations other than this can be easily used in our method. 
Thus, we created a matrix, $\W \in \mathcal{R}^{(d \times m)}$, where $d$ is the dimension of word representations and $m$ is the number of attributes. Each column of the matrix $\W$ represents \emph{word2vec} representations of an attribute. Some of the attributes, such as 
\emph{pointed nose} have multiple words. In such case, we average the representations of the component words to generate a vector. 
For a training image \x~with its ground truth vector, \y~which is filled with 0 (absence of attribute, we turn it to -1) or 1 (presence of attribute) we turn the 0s into -1s to negatee the representations of the attributes that are not present in the given image. Thus, our approach encodes the information which are present as well as the ones that are absent.
We then scale each of the attribute representation vectors by their corresponding labels and aggregate them to generate the final continuous value label of the image, $\z$. 
At training time, we learn the parameters to minimise the mean squared error loss on these continuous valued labels. The second part of Eqn.~\ref{eqn:obj_func} shows the continuous loss. 

\begin{equation}
    \mathcal{J(\theta)} = \alpha\frac{1}{N}\sum_{i=1}^{N} \sum_{j=1}^{m} -\y_{i,j}\log\y^{p}_{i, j} - (1- \y_{i,j})\log(1 - \y^{p}_{i, j}) + (1-\alpha) \frac{1}{N}\sum_{i=1}^{N} || \z_i - \z_{i}^{p}||^{2}_{2}   
\label{eqn:obj_func}
\end{equation}

We employed VGG16~\cite{simonyan2014very} as our baseline architecture similar to that of Sun et al.~\cite{sun2018bmvc}. We then generated the continuous valued labels to use them along with existing categorical labels. We used Stochastic Gradient Descent~\cite{bottou2010large} to learn the parameters that minimise the overall objective given in Eqn.~\ref{eqn:obj_func}.
The overall pipeline of our method is depicted in Fig.~\ref{fig:proposed_method}. Since the proposed method is independent of the architecture of the Deep CNN, it can be applied in any architecture other than VGG16.

\section{Experiments}
\label{experiments}

\subsection{Datasets}

\noindent \textbf{CelebA.} 
This is one of the largest and most widely used benchmarks for face attribute classification. This dataset consists of 200K annotated examples and is divided into training, validation and testing sets with sizes of 160K, 20K and 20K, respectively. There are 40 attributes in total.\\
\noindent \textbf{LFWA.} This is another one of the most challenging benchmarks, which is also widely used for face attribute classification.
There are more than 13K images, and each image is annotated with the presence or absence of 40 different attributes. This dataset is split into two equal halves as training and testing set. \\
\noindent \textbf{GloVe.} We use pre-trained word2vec representations to generate the new continuous labels. 
We use word representations with a dimension of $50$ to train the network efficiently. However, the word representations of other dimensions can be applied easily.

\subsection{Compared methods}
\noindent \textbf{No augmentation.}
We train the network with centre cropped pre-processed data without doing further augmentation in any of the spaces of the network, i'e input, intermediate and output. \\
\noindent \textbf{Geometric Augmentation (Geo. Aug.)}~\cite{krizhevsky2012imagenet}. This is a widely used and one of the most successful data augmentation techniques. We perform random cropping on the pre-processed data at 5 different locations (4 corners and 1 centre) followed by random flipping.\\
\noindent \textbf{Dropout}~\cite{srivastava2014dropout}. We randomly mask the neurons of the last 2 fully-connected layers to prevent the network from overfitting. \\
\noindent \textbf{Geo. Aug. + Dropout.}
Without loss of generality, geometric transformation of images and dropout of neurons are widely and successfully used to regularise the network. We train VGG16 initialised with vgg-face parameters as one of our baselines. 
\\
\noindent \textbf{DisturbLabel}~\cite{xie2016disturblabel}. This is a recent method, which is one of the closest label augmentation methods to ours. 
It proposes to randomly flip the categorical labels during training in addition to the above-mentioned network regularisation techniques.

\subsection{Pre-processing of Data and Implementation Details}
Images are frontalised and cropped to the size of $256 \times 256$. Before being fed into the networks, these images are cropped to the size of $224 \times 224$. 
To implement our baselines and proposed method, we use Keras~\cite{chollet2015keras} with Tensorflow backend. We conducted our experiments on a workstation with Intel i5-8500 3.0G, 32G memory, NVIDIA GTX1060 and NVIDIA GTX1080.   

\subsection{Evaluation metric} We compared our method with existing methods in both a quantitative and qualitative manner. We computed the mean accuracy over the 40 attributes to do quantitative evaluations, whereas we made graphical visualisations to compare the methods qualitatively.

\begin{table}[]
    \centering
    \resizebox{\textwidth}{!}{%
    \begin{tabular}{c|c|c}
     \hline 
     \textbf{Method} & \textbf{Mean Acc.} & \textbf{Remarks}\\
         \hline
     Baseline I & 85\%  & No Aug. \\ 
     \hline 
     Baseline II & 85.4\%  & Geo. Aug. \\ 
     \hline 
     Baseline III & 85.2\%  & Dropout \\ 
     \hline
     \textbf{AugLabel} & \textbf{85.8\%}  & Proposed method \\ 
     \hline
     \hline
     Baseline II + Baseline III & 85.6\% & Geo. Aug. + Dropout \\ 
     \hline
     Baseline II + AugLabel & \textbf{86.7\%} & Geo. Aug. + \textbf{AugLabel} \\ 
     \hline  
     Baseline II + Baseline III + AugLabel & \textbf{86.9\%} & Geo. Aug. + Dropout + \textbf{AugLabel} \\ 
     \hline  
     DisturbLabel~\cite{xie2016disturblabel}& 85.7\% & Geo. Aug. + Random flip label \\ 
     \hline  
     DisturbLabel~\cite{xie2016disturblabel} + Dropout & 85.5\% & Geo. Aug. + Dropout + Random flip label \\ 
     \hline 
     DisturbLabel~\cite{xie2016disturblabel} + AugLabel & \textbf{86.8\%} & Geo. Aug. + Random flip label + \textbf{AugLabel} \\ 
     \hline
     \hline  
     Face Tracer~\cite{kumar2008facetracer}  & 73.9\% &  \\
     \hline  
     PANDA~\cite{zhang2014panda} & 81\% &  \\
     \hline  
     LeNet+ANet~\cite{liu2015deep} & 83.9\% &  \\
     \hline  
     Walk and Learn~\cite{wang2016walk} & 86.6\% &  \\
     \hline 
     MCNN~\cite{hand2017attributes} & 86.3\%  & \\
     \hline 
     Kalayeh et al.~\cite{kalayeh2017improving} & \textbf{87.1\%} & \\
     \hline 
     Sun et al.~\cite{sun2018bmvc}& \textbf{87.1\%} & \\
     \hline  
    \end{tabular}}
    \caption{Performance comparison on LFWA with existing arts}
    \label{tab:exp_compare_lfwa_arts}
\end{table}

\subsection{Quantitative Evaluation.}
We evaluated the proposed methods on two challenging face attribute classification benchmarks, namely LFWA and CelebA. In this section, we present our results and compare them with existing arts and several baselines in detail.\\ 

\noindent \textbf{Evaluation on LFWA.} Tab.~\ref{tab:exp_compare_lfwa_arts} tabulates the performance of several baselines and existing arts. We have split the experiments and existing arts into three groups. The top group shows the performance of different baselines. From this group we can clearly see that the regularisation techniques, Geometric transformation(Geo. Aug.), Drop out and the proposed method, AugLabel, effectively generalise the model.
Geo Aug and Drop out alone improve the performance over baseline (no augmentation) by +0.4\% and +0.2\%, respectively. 
Our proposed method, AugLabel, improves the performance of the model by +0.8\% over the baseline with no augmentation. This shows that our method is superior to the existing most successful and popular generalisation techniques.In addition to this, we also carefully evaluated the \textbf{combinatorial effect} of these generalisation techniques on the performance of CNNs.

The middle block on Tab.~\ref{tab:exp_compare_lfwa_arts} shows the performance on LFWA when the augmentation techniques are applied in different combinations. 
From this group we can see that when AugLabel is applied together with Geo. Aug.+Dropout, it surpasses the most popular method to regularise the network by +1.9\%. 

In addition to this, we also compare our method with one of the recent methods to augment in label space, DisturbLabel. The application of Dropout on DisturbLabel slightly lowered the performance of DisturbLabel. When we applied our approach on DisturbLabel, we observe a boost in the performance by +1.1\%. This demonstrates that the  proposed method is simple, effective and complementary to existing methods. The last block of Tab.~\ref{tab:exp_compare_lfwa_arts} shows the performance of existing arts on LFWA. From this block, we can see that the performance of the proposed method is competitive to the existing state-of-the-art methods~\cite{sun2018bmvc,kalayeh2017improving}. However, our method is not directly comparable to these methods as Sun et al.~\cite{sun2018bmvc} has more than $40\times$ of the parameters than our method does. Similarly, ~\cite{kalayeh2017improving} relies on semantic segmentation of attributes, which is comparatively more difficult to compute than unsupervised \emph{word2vec} representations. Finally, our method is a generic regularisation technique and can be easily applied to existing arts. One of the arts from the third block, which is comparable to our method is MCNN~\cite{hand2017attributes}. Our method outperforms this method by +0.6\%. \\

\noindent \textbf{Evaluation on CelebA.} This is another challenging standard benchmark for face attribute classification. Without loss of generality, Geo. Aug.+Dropout is the most successful and widely used combination of regularisation techniques.
Hence, we took this as the baseline and applied our proposed method. We can see an improvement over the baseline by +0.3\% attaining 91.2\%. Please refer to Tab.~\ref{tab:exp_compare_celeba_arts} for the details. In addition, our method is competitive to existing state-of-the-art methods. As stated before, our method is not directly comparable to the methods: AFFACT~\cite{gunther2017affact} and of Sun et al.~\cite{sun2018bmvc} as these methods rely on multiple number of CNNs. Moreover, our method is generic and complementary in nature and can be applied in any architecture of CNNs.\\

\begin{figure}
    \centering
    \includegraphics[width=0.6\linewidth]{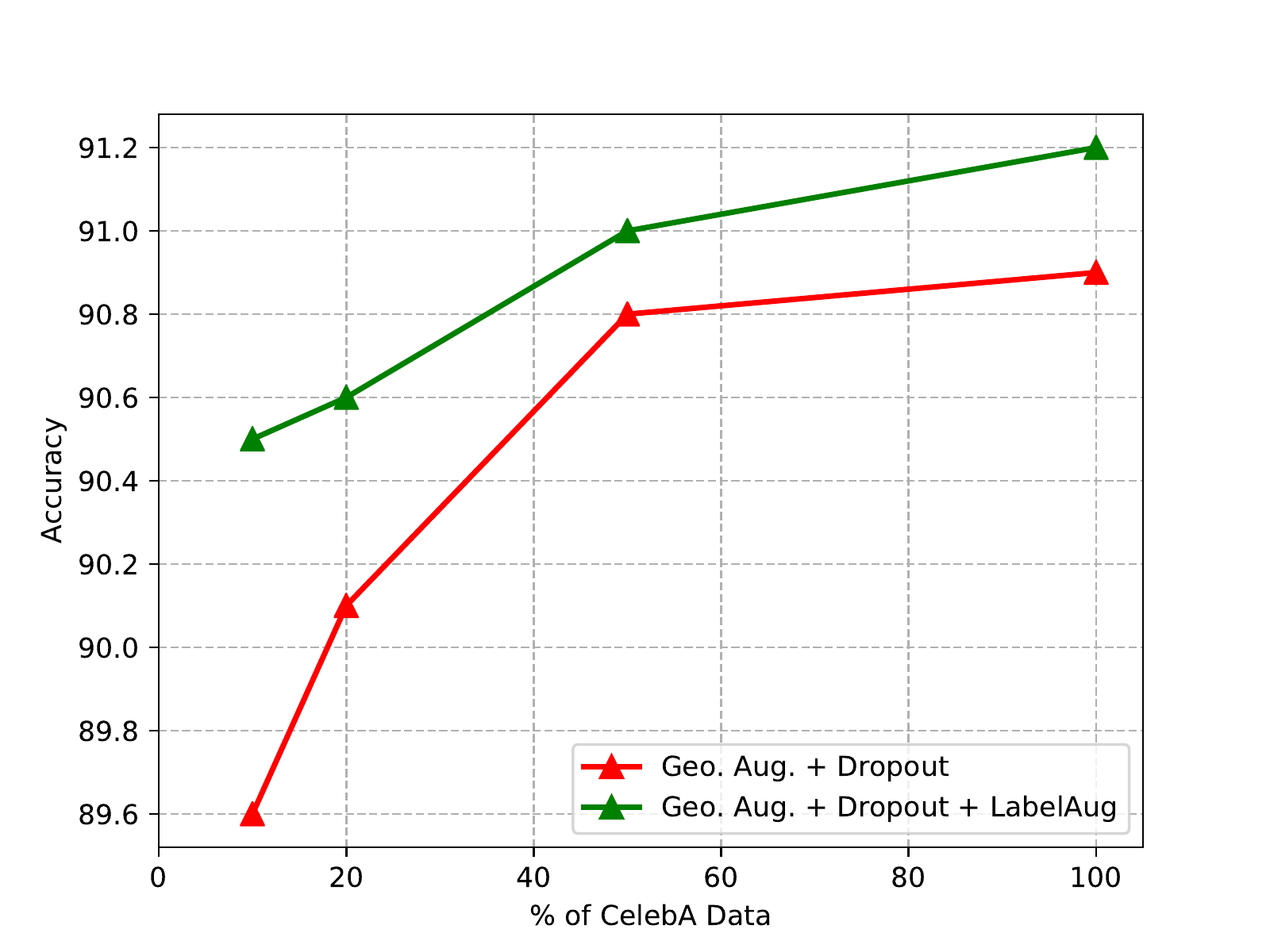}
    \caption{Performance comparison on CelebA with different size of training examples.
    Using the augmented labels reduces the need of annotated real data up to 50\%. }
    \label{fig:celeba_size}
    \vspace{-0.5cm}
\end{figure}

\noindent \textbf{Reduction on training data set.} To check the impact of our method on small scale to moderate size training data set, we randomly 
selected  10\%, 20\% and 50\% of the training data from CelebA. We then applied our method over the most competitive generalisation techniques 
i'e Geo. Aug. + Dropout. Fig.~\ref{fig:celeba_size} compares the performance of the baseline and the proposed method. From the Fig., we can see that the performance of the baseline obtained by 20\% (90.1\%) of the training data is obtained by our method by only using 10\% of the training data ($90.2\%$). Similarly, performance on full data set by the baseline method (90.9\%) is obtained by our method just by using half of the full data set (91.0\%). 

\begin{table}[]
    \centering
    \begin{tabular}{c|c}
     \hline 
     \textbf{Method} & \textbf{Mean Acc.} \\
     \hline
     Baseline (Geo. Aug. + Dropout) & 90.9\%  \\ 
     \hline  
     Proposed (Geo. Aug. + Dropout + \textbf{AugLabel}) & \textbf{91.2\%} \\
     \hline  
     \hline  
     Face Tracer~\cite{kumar2008facetracer} & 81.1\%   \\
     \hline  
     LeNet+ANet~\cite{liu2015deep}& 87.3\%   \\
     \hline  
     MOON~\cite{rudd2016moon} & 90.9\%   \\
     \hline  
     Walk and Learn~\cite{wang2016walk} & 88.7\%   \\
     \hline  
     MCNN~\cite{hand2017attributes} & 91.2\%  \\
     \hline  
     AFFACT~\cite{gunther2017affact} & 91.5\%   \\
     \hline 
     Kalayeh et al.~\cite{kalayeh2017improving}  & 91.2\%  \\
     \hline 
     Sun et al.~\cite{sun2018bmvc} & \textbf{91.6\%}  \\
     \hline  
    \end{tabular}
    \caption{Performance comparison on CelebA with existing arts}
    \label{tab:exp_compare_celeba_arts}
    \vspace{-0.5cm}
\end{table}

\subsection{Qualitative Evaluations}

\begin{figure}
    \centering
    \includegraphics[width=0.9\textwidth, trim={0cm 0cm 0cm 0cm},  clip]{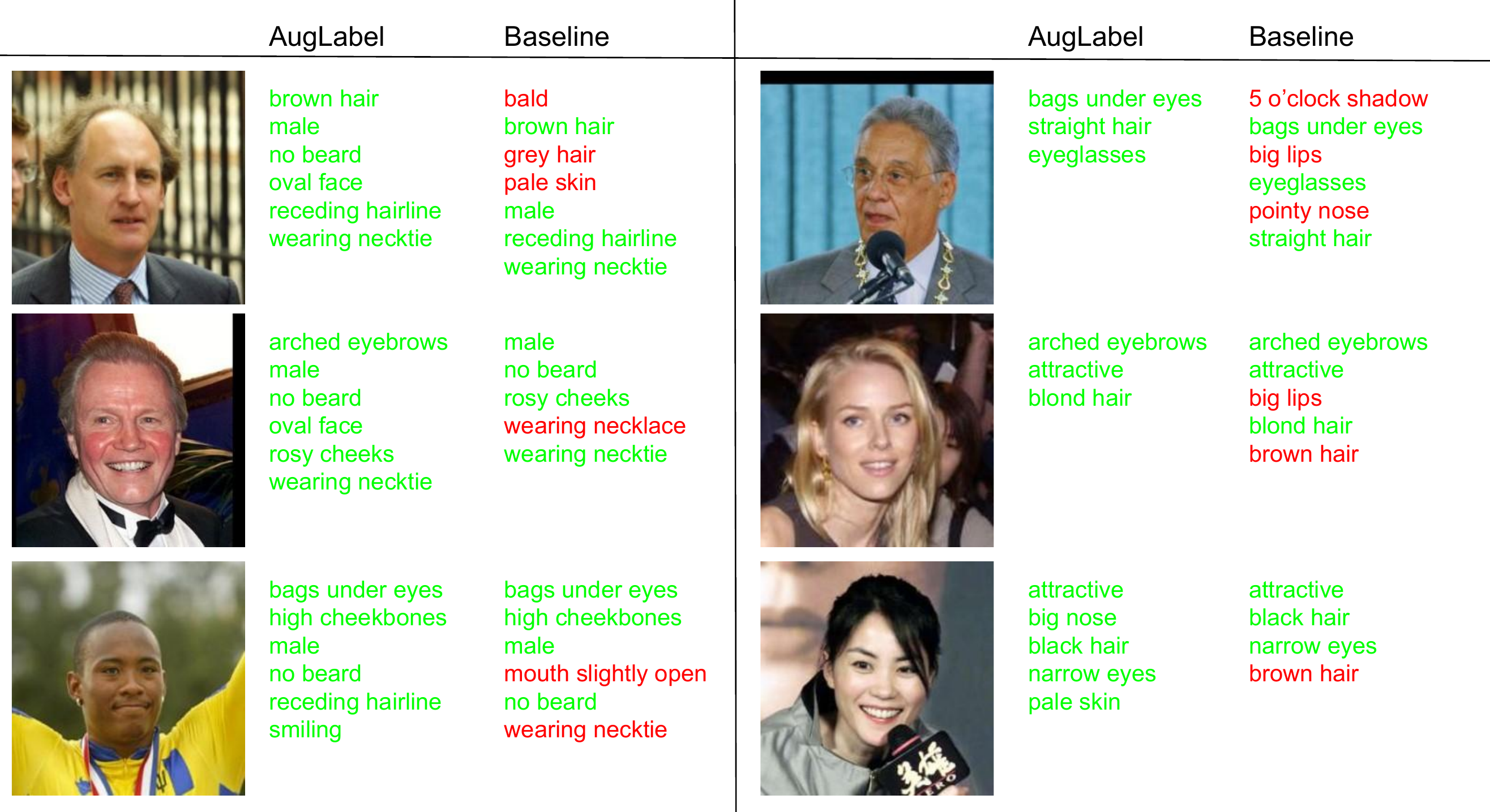}
    \caption{
    The listed attributes are a subset of the attributes predicted by the corresponding model on CelebA. The green colour indicates that the predicted attribute is present in the ground truth, while the red ones are absent (best viewed in colour). 
   }
    \label{fig:qual_viz}
\end{figure}

Fig.~\ref{fig:qual_viz} shows qualitative visualisations of some of the predictions made by the baseline (Geo. Aug. + Dropout) and the proposed method.
Green coloured attributes are attributes that correctly classified as present whereas the red coloured attributes are wrongly classified and are in fact absent according to the ground truth. In the Fig., we can observe that some of the difficult and confusing attributes are classified correctly by the proposed method. For instance, the top left image in the Fig., the person is partly \emph{bald}, and the baseline predicted it as \emph{bald}, \emph{grey haired} and \emph{brown haired} whereas our method could make the correct decision.
Similarly, the image on the left middle is classified wrongly by the baseline since the tie knot looks like a necklace, but the proposed method was robust to it. 
\emph{Mouth slightly open} on third left and \emph{brown hair} on second right Figs. are quite difficult to classify even for human. In these two cases too, baseline failed whereas the proposed method succeeded.

\section{Conclusions}
\label{conclusions}
In this paper, we present AugLabel; a simple, efficient and effective label augmentation technique to generalise the parameters of CNNs for face attribute classification. To this end, we propose to use \emph{word2vec} representations of words to generate continuous value filled labels and augment existing categorical labels with them. Our extensive experiments on two challenging benchmarks demonstrate that our method is superior to contemporary regularisation techniques. In addition, our method is complementary in nature and applying along with other existing methods further improves the performance of the existing methods.

As a future work, we plan to explore the methods to re-weight the contribution of the component attributes instead of giving uniform weights to generate final continuous value filled vectors. We also plan to explore Graph CNN to learn such relationships between the component vectors. 
\subsection{Acknowledgements}
This work is partly supported by EPSRC FACER2VM project.
\printbibliography

\end{document}